\def\year{2020}\relax
\documentclass[letterpaper]{article} 
\usepackage{aaai20}  
\usepackage{times}  
\usepackage{helvet} 
\usepackage{courier}  
\usepackage[hyphens]{url}  
\usepackage{graphicx} 
\usepackage{graphicx}  
\usepackage{amsmath}
\usepackage{amssymb}
\usepackage{array}
\usepackage{multirow}
\usepackage{booktabs}
\usepackage{subfigure}
\title{Tree-Structured Policy based Progressive Reinforcement Learning \\for Temporally Language Grounding in Video}
\author{Jie Wu\textsuperscript{\rm 1}, Guanbin Li\textsuperscript{\rm 1}\thanks{Corresponding author is Guanbin Li. This work was supported in part by the National Key Research and Development Program of China under Grant No. 2018YFC0830103, in part by the National Natural Science Foundation of China under Grant No.61976250 and No.61876177, in part by the National High Level Talents Special Support Plan~(Ten Thousand Talents Program). This work was also supported by sponsored by CCF-Tencent Open Research Fund~(CCF-Tencent IAGR20190106).}, Si Liu\textsuperscript{\rm 2}, Liang Lin\textsuperscript{\rm 1,3}
\\ \textsuperscript{\rm 1} Sun Yat-sen University \textsuperscript{\rm 2} Beihang University, \textsuperscript{\rm 3} DarkMatter AI Research. \\
wujie23@mail2.sysu.edu.cn, liguanbin@mail.sysu.edu.cn, liusi@buaa.edu.cn, linliang@ieee.org
}

\begin{document}

\maketitle

\begin{abstract}
Temporally language grounding in untrimmed videos is a newly-raised task in video understanding.
Most of the existing methods suffer from inferior efficiency, lacking interpretability, and deviating from the human perception mechanism.
Inspired by human's coarse-to-fine decision-making paradigm, we formulate a novel Tree-Structured Policy based Progressive Reinforcement Learning (TSP-PRL) framework to sequentially regulate the temporal boundary by an iterative refinement process.
The semantic concepts are explicitly represented as the branches in the policy, which contributes to efficiently decomposing complex policies into an interpretable primitive action.
Progressive reinforcement learning provides correct credit assignment via two task-oriented rewards that encourage mutual promotion within the tree-structured policy.
We extensively evaluate TSP-PRL on the Charades-STA and ActivityNet datasets, and experimental results show that TSP-PRL achieves competitive performance over existing state-of-the-art methods.
\end{abstract}

\section{Introduction}
We focus on the task of temporally language grounding in a video, whose goal is to determine the temporal boundary of the segments in the untrimmed video that corresponds to the given sentence statement.
Most of the existing competitive approaches \cite{anne2017localizing,gao2017tall,liu2018attentive,ge2019mac,xu2019multilevel} are based on extensive temporal sliding windows to slide over the entire video or rank all possible clip-sentence pairs to obtain the grounding results. However, these sliding window based methods suffer from inferior efficiency and deviate from the human perception mechanism.
When humans locate an interval window associated with a sentence description in a video, they tend to assume an initial temporal interval first, and achieve precise time boundary localization through cross-modal semantic matching analysis and sequential boundary adjustment~(e.g., scaling or shifting).

\begin{figure*}[t]
    \centering
    \includegraphics[width=0.95\linewidth]{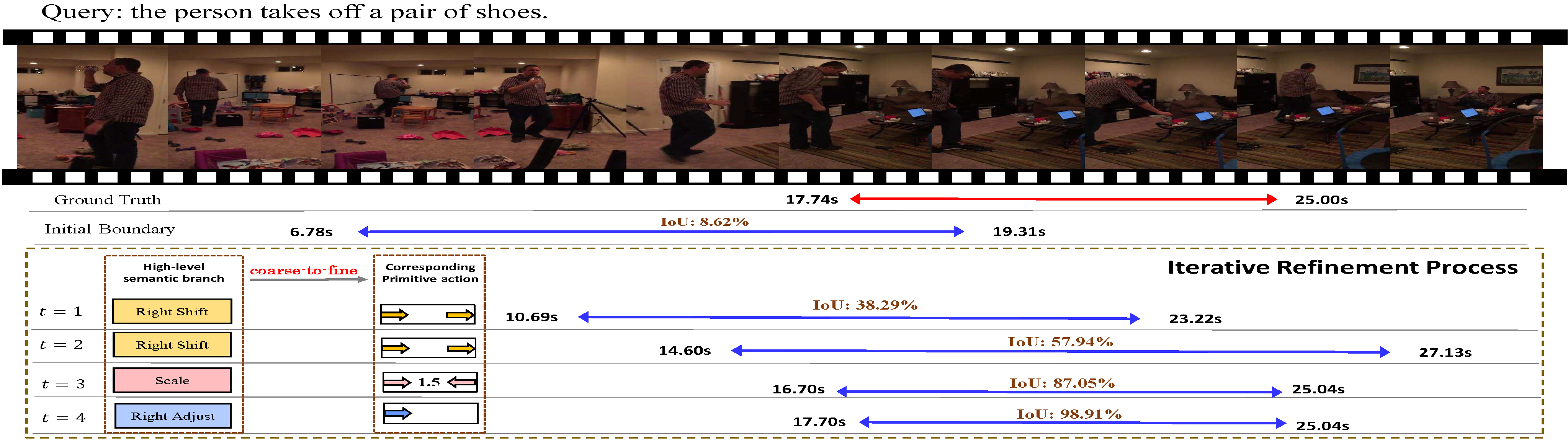}
    \caption{An example showing how TSP-PRL addresses the task in an iterative refinement manner. A more interpretable primitive action is proposed by the tree-structured policy, which consists of root policy and leaf policy to imitate human's coarse-to-fine decision-making scheme.}
    \label{fig:introduction}
\end{figure*}

Looking deep into human's thinking paradigm \cite{mancas2016human}, people usually deduce a coarse-to-fine deliberation process to render a more reasonable and interpretable decision in daily life.
Namely, people will first roughly determine the selection range before making a decision, then choose the best one among the coarse alternatives. This top-down coarse-to-fine deliberation has been explored in the task of machine translation, text summarization and so on \cite{NIPS2017_6775}.
Intuitively, embedding this mode of thinking into our task can efficiently decompose complex action policies, reduce the number of search steps while increasing the search space, and obtain more impressive results in a more reasonable way.
To this end, we formulate a Tree-Structured Policy based Progressive Reinforcement Learning framework (TSP-PRL) to imitate human's coarse-to-fine decision-making scheme.
The tree-structured policy in TSP-PRL consists of root policy and leaf policy, which respectively correspond to the process of coarse and fine decision-making stage. And a more reasonable primitive action is proposed via these two-stages selection.
The primitive actions are divided into five classes related to semantic concepts according to the moving distance and directions: scale variation, markedly left shift, markedly right shift, marginally left adjustment and marginally right adjustment. The above semantic concepts are explicitly represented as the branches into the tree-structured policy, which contributes to efficiently decomposing complex policies into an interpretable primitive action.
In the reasoning stage, the root policy first roughly estimates the high-level semantic branch that can reduce the semantic gap to the most extent. Then the leaf policy reasons a refined primitive action based on the selected branch to optimize the boundary.
We depict an example of how TSP-PRL addresses the task in Figure \ref{fig:introduction}.
As can be seen in the figure, the agent first markedly right shift the boundary to eliminate the semantic gap. Then it resorts to scale contraction and marginally adjustment to obtain an accurate boundary.

The tree-structured policy is optimized via progressive reinforcement learning, which determines the selected single policy (root policy or leaf policy) in the current iteration while stabilizing the training process.
The task-oriented reward settings in PRL manages to provide correct credit assignment and optimize
the root policy and leaf policy mutually and progressively. Concretely, the external environment provides rewards for each leaf strategy and the root strategy does not interact directly with the environment. PRL measures the reward for the root policy from two items:
1) the intrinsic reward for the selection of high-level semantic branch; 2) the extrinsic reward that reflects how the subsequent action executed by the selected semantic branch influences the environment.

Extensive experiments on Charades-STA \cite{sigurdsson2016hollywood,gao2017tall} and ActivityNet \cite{krishna2017dense} datasets prove that TSP-PRL achieves competitive performance over existing leading and baseline methods on both datasets.
The experimental results also demonstrate that the proposed approach can (i) efficiently improve the ability to discover complex policies which can hardly be learned by flat policy; (ii) provide more comprehensive assessment and appropriate credit assignment to optimize the tree-structured policy progressively; and (iii) determine a more accurate stop signal at an iterative process. The source code as well as the trained models have been released at~\emph{https://github.com/WuJie1010/TSP-PRL}.

\section{Related work}
\begin{figure*}
\centering
\includegraphics[width=0.95\linewidth]{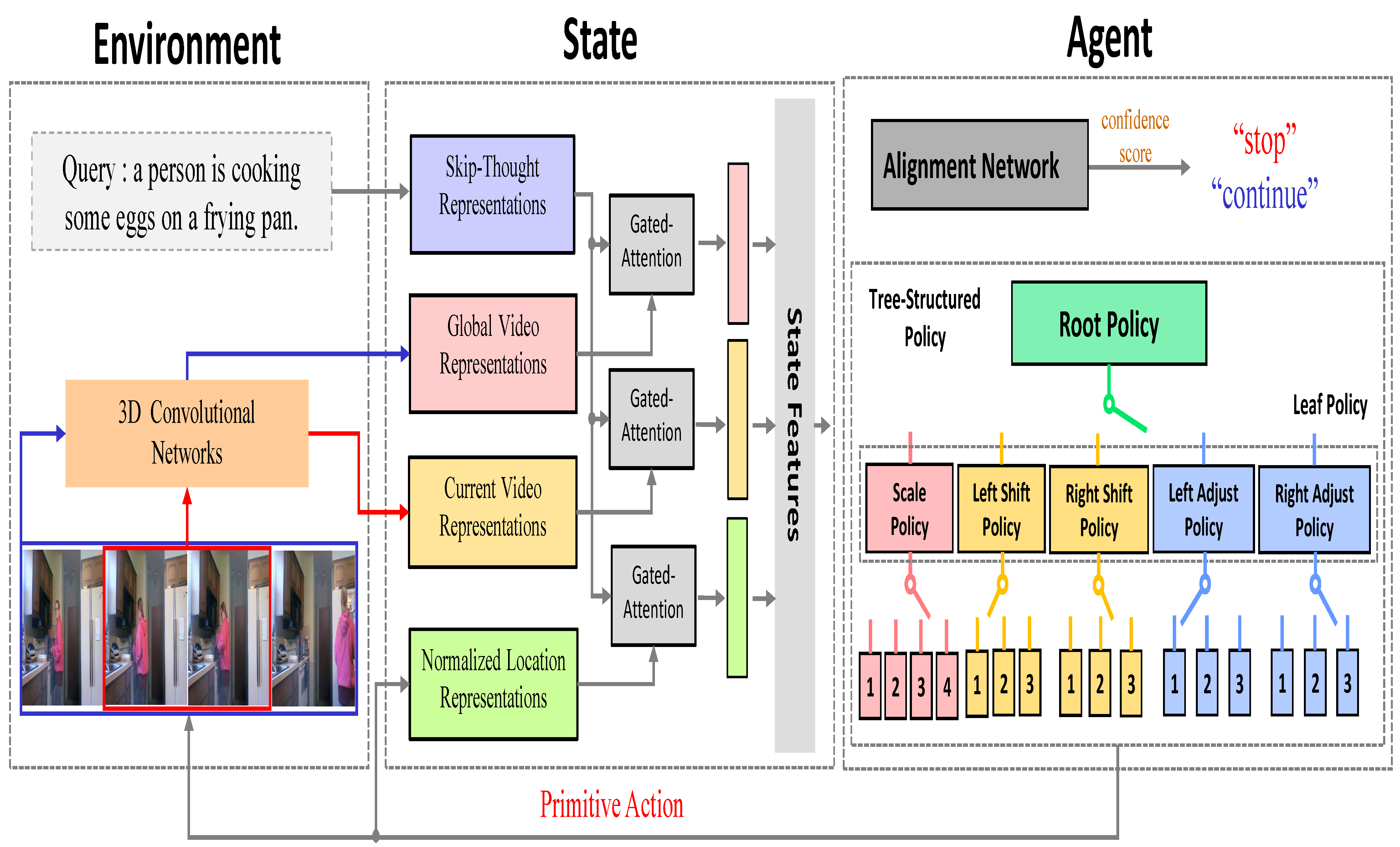}
\caption{The overall pipeline of the proposed TSP-PRL framework. The agent receives the state from the environment (video clips) and estimates a primitive action via tree-structured policy. The action selection is depicted by a switch $\multimap$ over the interface $\bot$ in the tree-structured policy. The alignment network will predict a confidence score to determine when to stop.}
\label{fig:model}
\end{figure*}
\noindent\textbf{Temporally Language Grounding in Video}.
Temporally language grounding in the video is a challenging task which requires both language and video understanding and needs to model the fine-grained interactions between the verbal and visual modalities.
Gao \emph{et al.} \cite{gao2017tall} proposed a cross-modal temporal regression localizer (CTRL) to jointly model language query and video clips, which adopts sliding windows over the entire video to obtain the grounding results.
Hendricks \emph{et al.}\cite{anne2017localizing} designed a moment context network (MCN) to measure the distance between visual features and sentence embedding in a shared space, ranking all possible clip-sentence pairs to locate the best segments.
However, the above approaches are either inefficient or inflexible since they carry out overlapping sliding window matching or exhaustive search.
Chen \emph{et al.} \cite{chen2018temporally} designed a dynamic single-stream deep architecture to incorporate the evolving fine-grained frame-by-word interactions across video-sentence modalities. This model performs efficiently, which only needs to process the video sequence in one single pass.
Zhang \emph{et al.} \cite{zhang2019man} exploited graph-structured moment relations to model temporal structures and improve moment representation explicitly.
He \emph{et al.} \cite{he2019read} first introduced the reinforcement learning paradigm into this task and treated it as a sequential decision-making task.
Inspired by human's coarse-to-fine decision-making paradigm, we construct a tree-structured policy to reason a series of interpretable actions and regulate the boundary in an iterative refinement manner.

\noindent\textbf{Reinforcement Learning}.
Recently, reinforcement learning (RL) technique \cite{williams1992simple} has been successfully popularized to learn task-specific policies in various image/video-based AI tasks.
These tasks can be generally formulated as a sequential process that executes a series of actions to finish the corresponding objective.
In the task of multi-label image recognition, Chen \emph{et al.} \cite{chen2018recurrent} proposed a recurrent attentional reinforcement learning method to iteratively discover a sequence of attentional and informative regions.
Shi \emph{et al.}\cite{shi2019face} implemented deep reinforcement learning and developed a novel attention-aware face hallucination framework to generate a high-resolution face image from a low-resolution input.
Wu \emph{et al.} \cite{wu2019concrete} designed a new content sensitive and global discriminative reward function to encourage generating more concrete and discriminative image descriptions.
In the video domain, RL has been widely used in temporal action localization \cite{yeung2016end} and video recognition \cite{wu2019multi}.
In this paper, we design a progressive RL approach to train the tree-structured policy, and the task-oriented reward settings contribute to optimizing the root policy and leaf policy mutually and stably.

\section{The Proposed Approach}
\subsection{Markov Decision Process Formulation}
In this work, we cast the temporally language grounding task as a Markov Decision Process (MDP), which is represented
by states $s \in \mathcal{S}$, action tuple $\left\langle a^r, a^l\right\rangle$, and transition function $\mathcal{T} :(s, \left\langle a^r, a^l\right\rangle) \rightarrow s^{\prime}$.
$a^r$ and $a^l$ denote the action proposed by root policy and leaf policy, respectively. The overall pipeline of the proposed Tree-Structured Policy based Progressive Reinforcement Learning (TSP-PRL) framework is depicted in Figure \ref{fig:model}.

\noindent\textbf{State}.
A video is firstly decomposed into consecutive video units ~\cite{gao2017tall} and each video unit is used to extract unit-level feature through the feature extractor $\varphi_v$ \cite{tran2015learning,wang2016temporal}.
Then the model resorts to uniformly sampling strategy to extract ten unit-level features from the entire video, which are concatenated as the global video representation $V^g$. For the sentence query $L$, the skip-thought encoder $\varphi_s$ \cite{kiros2015skip} is utilized to generate the language representation $E = \varphi_s(L) $. When the agent interacts with the environment, the above features are retained.
At each time step, the action executed by the agent will change the boundary and obtain a refined video clip. The model samples ten unit-level features inside the boundary and concatenate these features as the current video feature $V^c_{t-1}$, $t=1,2,...,T_{max}$. We explicitly involve the normalized boundary $L_{t-1}=[l^s_{t-1},l^e_{t-1}]$ into the state feature \cite{he2019read}, where $l^s_{t-1}$ and $l^e_{t-1}$ denote the start point and end point respectively. Then the gated-attention \cite{chaplot2018gated} mechanism is applied to gain multi-modal fusion representation of verbal and visual modalities:
\begin{equation}
\begin{split}
A^{EG}_t &= \sigma (E) \odot V^g, \quad A^{EC}_t = \sigma (E) \odot  V^c_{t-1}, \\
&\quad A^{EL}_t = \sigma (E) \odot L_{t-1},
\end{split}
\end{equation}
where $\sigma$ denotes the sigmoid activation function and $\odot$ is the Hadamard product.
The above gated attention features are concatenated and fed into a fully-connected layer $\phi$ to obtain the state representation $s_t$:
\begin{equation}
s_t = \phi (A^{EG}_t, A^{EC}_t, A^{EL}_t)
\label{eq1}
\end{equation}

An additional GRU \cite{cho2014learning} layer is adopted to process the state features before feeding them into the tree-structured policy, which manages to develop high-level temporal abstractions and lead to a more generalizable model.

\noindent\textbf{Hierarchical Action Space}.
In our work, the boundary movement is based on the clip-level and each boundary consists of a series of video clips.
All primitive actions can be divided into five classes related to semantic concepts according to the moving distance and directions, which results in a hierarchical action space on the whole.
These semantic concepts are explicitly represented as the branches into the tree-structured policy, resulting in five high-level semantic branches to contain all primitive actions in this task: scale variation, markedly left shift, markedly right shift, marginally left adjustment and marginally right adjustment.
i) The scale variation branch contains four primitive actions: extending/shortening $\xi$ times w.r.t center point. $\xi$ is set to 1.2 or 1.5;
2) Three actions are included in the markedly left shift branch: shifting start point/end point/start \& end point backward $\nu$. $\nu$ is fixed to $N/10$, where $N$ denotes the number of the clip of the entire video;
3) The actions in the markedly right shift branch is symmetry with the markedly left shift: shifting start point/end point/start \& end point forward $\nu$;
4) Except for the moving scale, the actions in the marginally left adjustment branch is similar to the markedly left shift branch: shifting start point/end point/start \& end point backward $Z$ frame; The size of $Z$ is constrained by the video lengths;
5) The marginally right adjustment branch also involves three primitive actions: shifting start point/end point/start \& end point forward $Z$ frame.

\begin{figure*}
    \centering
    \includegraphics[width=0.98\linewidth]{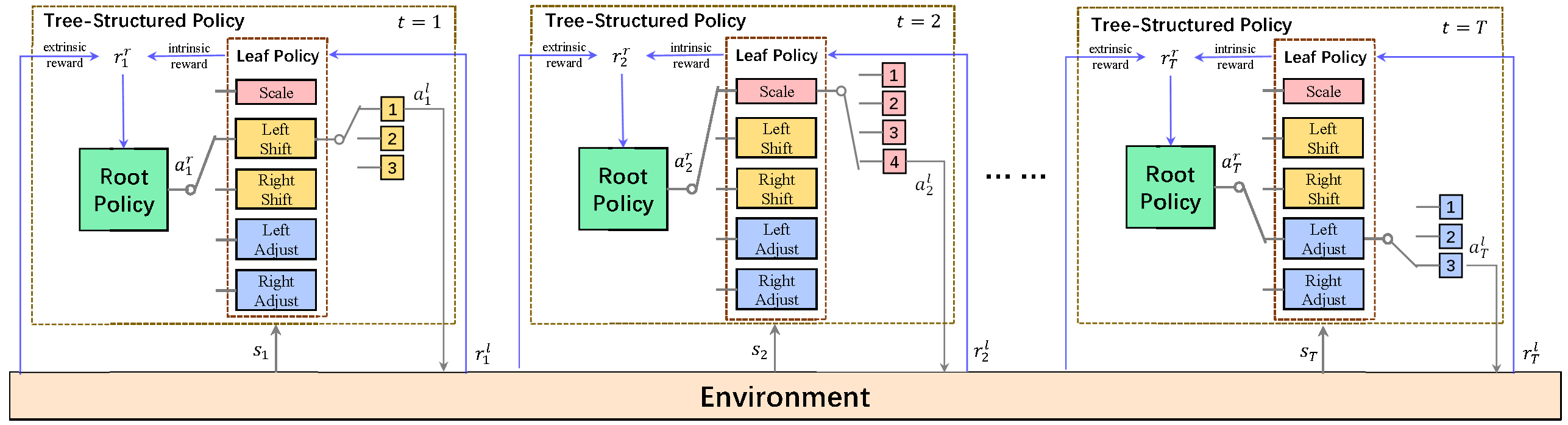}
    \caption{An illustration of how the tree-structured policy works iteratively. The solid blue line represents how the root reward and leaf reward are obtained from the proposed framework.}
    \label{fig:timestep}
\end{figure*}

\noindent\textbf{Tree-Structured Policy}.
One of our key ideas is that the agent needs to understand the environmental state well and reason a more interpretable primitive action. Inspired by human's coarse-to-fine decision-making paradigm, we design a tree-structured policy to decompose complex action policies and propose a more reasonable primitive action via two-stages selection, instead of using a flat policy that maps the state feature to action directly \cite{he2019read}.
 As shown in the right half of Figure \ref{fig:model}, the tree-structured policy consists of a root policy and a leaf policy at each time step. The root policy $\pi^{r}(a^r_t|s_t)$ decides which semantic branch will be primarily relied on. The leaf policy $\pi^{l}(a^l_t|s_t, a^r_t)$ consists of five sub-policies, which corresponds to five high-level semantic branches.
The selected semantic branch will reason a refined primitive action via the corresponding sub-policy.
The root policy aims to learn to invoke the correct sub-policy from the leaf policy in the following different situations:
(1) The scaling policy should be selected when the scale of predicted boundary is quite mismatched with the ground-truth boundary; (2) When the predicted boundary is far from the ground-truth boundary, the agent should execute the left or right shift policy; (3) The primitive action should be sampled from the left or right adjust policy when most of the two boundaries intersect but with some deviation.
At each time step, the tree-structured policy first samples $a^r_t$ from root policy $\pi^{r}$ to decide the semantic branch:
\begin{equation}
a^r_t \sim \pi^{r}(a^r_t|s_t).
\label{eq2}
\end{equation}
And a primitive action is sampled from the leaf policy $\pi^{l}$ related to the selected semantic branch:
\begin{equation}
a^l_{t} \sim \pi^{l}(a^l_{t} | s_{t}, a^r_t).
\label{eq3}
\end{equation}

\subsection{Tree-Structured Policy based Progressive Reinforcement Learning}

\noindent\textbf{Rewards}.
Temporal IoU is adopted to measure the alignment degree between the predicted boundary $[l^s$, $l^e]$ and ground-truth boundary $[g^s$, $g^e]$:
\begin{equation}
U_{t}=\frac{\min \left(g^{e}, l_{t}^{e}\right)-\max \left(g^{s}, l_{t}^{s}\right)}{\max \left(g_{e}, l_{t}^{e}\right)-\min \left(g^{s}, l^{s}_{t}\right)}.
\label{eq10}
\end{equation}

The reward setting for this task should provide correct credit assignment, encouraging the agent to take fewer steps to obtain accurate grounding results. We define two task-oriented reward functions to select an accurate high-level semantic branch and the corresponding primitive action, respectively. The first reward $r^l_t$ is the leaf reward, which reveals the influence of the primitive actions $a^l_{t}$ to the current environment. It can be directly obtained in the environment through temporal IoU. We explicitly provide higher leaf reward when the primitive action attempts to obtain better grounding results and the temporal IoU is higher than 0.5:
\begin{equation}
\begin{split}
r^l_t =
\begin{cases}
\zeta + U_{t} & \text{$U_{t}>U_{t-1}$; $U_{t} > 0.5$ }\\
\zeta & \text{$U_{t}>U_{t-1}$; $U_{t} \leq 0.5$ }\\
-\zeta /10  & \text{$U_{t-1} \geq U_{t} \geq 0$}\\
-\zeta & \text{otherwise}
\end{cases},
\end{split}
\label{eq11}
\end{equation}
where $\zeta$ is a factor that determines the degree of reward.

The second reward is the root reward $r^r_t$, which should be determined deliberately since the action executed by root policy does not interact with the environment directly.
To provide  comprehensive assessment and correct credit assignment, $r^r_t$ is defined to include two items: 1) the intrinsic reward term that represents the direct impact of $a^r_t$ for semantic branch selection and 2) the extrinsic reward term reflects the indirect influence of the subsequent primitive action executed by the selected branch for the environment.
In order to estimate how well the root policy chooses the high-level semantic branch, the model traverses through all possible branches and reasons the corresponding primitive actions to the environment, which results in five different IoU. The max IoU among these five IoU is defined as $U^{\operatorname{max}}_{t}$. Then the root reward $r^r_t$ is designed as follow:
\begin{equation}
\begin{split}
r^r_t =
\begin{cases}
\underbrace{\zeta}_{\text{intrinsic reward item}}  + \underbrace{U_{t} - U_{t-1}}_{\text{extrinsic reward item}}& \text{$U_{t}=U^{\operatorname{max}}_{t} $}\\
\underbrace{U_t - U^{\operatorname{max}}_{t}}_{\text{intrinsic reward item}} + \underbrace{U_{t} - U_{t-1}}_{\text{extrinsic reward item}}& \text{otherwise}
\end{cases},
\end{split}
\label{eq12}
\end{equation}
where $U_{0}$ denotes the temporal IoU between initial boundary and the ground-truth boundary. The diagram of how the root reward and leaf reward are obtained in the framework is depicted in Figure \ref{fig:timestep}.

\noindent\textbf{Progressive Reinforcement Learning}.
Progressive Reinforcement Learning (PRL) is designed on the basis of the advantage actor-critic (A2C) \cite{sutton2018reinforcement} algorithm to optimize the overall framework. Policy function $\pi^{r}(a^r_{t}|s_{t})$ and $\pi^{l}(a^l_{t} | s_{t},a^r_t)$ estimate the probability distribution over possible actions in the corresponding action space. These two policies are separate and each is equipped with a value approximator $V^{r}(s_{t})$ and $V^{l}(s_{t},a^r_t)$, which is designed to compute a scalar estimate of reward for the corresponding policy.

Starting from the initial boundary, the agent invokes the tree-structured policy iteratively in the interaction process. We depict how the tree-structured policy works iteratively in Figure \ref{fig:timestep}. From the figure, we can observe that the agent samples actions from root policy and leaf policy consecutively at each time step. The action will trigger a new state, which is fed into the tree-structured policy to execute the next actions.
Given a trajectory in an episode $\Gamma = \{\langle s_t, \pi^{r}(\cdot | s_t), a^r_t, r^r_t, \pi^{l}(\cdot | s_t, a^r_t), a^l_t, r^l_t \rangle, t = \{1, \cdots, T_{max}\}\}$, PRL algorithm maximizes the objective of root policy $\mathcal{L}_{root}(\theta_{\pi^{r}})$ and leaf policy $ \mathcal{L}_{leaf}(\theta_{\pi^{l}})$:
\begin{footnotesize}
\begin{equation}
\begin{split}
 \mathcal{L}_{root}(\theta_{\pi^{r}})= &-\frac{1}{M}\sum_{m=1}^{M} \sum_{t=1}^{T_{max}} [\mathrm{log} \pi^{r}(a^r_t|s_t)(R^r_t-V^{r}(s_{t}))\\
&+ \alpha H(\pi^{r}(a^r_t|s_t))],
\end{split}
\label{eq4}
\end{equation}
\end{footnotesize}
\begin{footnotesize}
\begin{equation}
\begin{split}
 \mathcal{L}_{leaf}(\theta_{\pi^{l}})= &-\frac{1}{M}\sum_{m=1}^{M} \sum_{t=1}^{T_{max}}[\mathrm{log} \pi^{l}(a^l_t|s_t, a^r_t)( R^l_t-V^{l}(s_{t}, a^r_t)) \\
 &+ \alpha H(\pi^{l}(a^l_t|s_t,a^r_t))],
\end{split}
\label{eq5}
\end{equation}
\end{footnotesize}
where $M$ denotes the size of a mini-batch and $T_{max}$ is the max time step in an episode. $R^r_t-V^{r}(s_{t})$ and $R^l_t-V^{l}(s_{t},a^r_t)$ denote the advantage functions in the A2C setting. $H()$ is the entropy of policy networks and the hyper-parameters $\alpha$ controls the strength of entropy regularization term, which is introduced to increase the diversity of actions.  $\theta_{\pi^{r}}$ and $\theta_{\pi^{l}}$ are the parameters of the policy networks. Here, the model only back-propagates the gradient for the selected sub-policy in leaf policy. The reward of the following-up steps should be traced back to the current step since it is a sequential decision-making problem. The accumulated root reward function $R^r_t$ is computed as follows:
\begin{equation}
\begin{split}
R^r_t =
\begin{cases}
r^r_t + \gamma  V^{r}(s_{t}) & \text{$t = T_{max}$}\\
r^r_t + \gamma  R^r_{t+1} & \text{$t = 1,2,...,T_{max}-1 $}
\end{cases},
\end{split}
\label{eq6}
\end{equation}
where $\gamma$ is a constant discount factor and the accumulated leaf reward $R^l_t$ is obtained in a similar way.
In order to optimize the value network to provide an estimation of the expected sum of rewards, we minimize the squared difference between the accumulated reward and the estimated value, and minimize the value loss:
\begin{equation}
\begin{split}
& \mathcal{L}_{root}(\theta_{V^{r}})=  \frac{1}{M}\sum_{m=1}^{M} \sum_{t=1}^{T_{max}}(R^r_t-V^{r}(s_{t}))^2, \\
 & \mathcal{L}_{leaf}(\theta_{V^{l}})= \frac{1}{M}\sum_{m=1}^{M} \sum_{t=1}^{T_{max}} (R^l_t-V^{l}(s_{t}, a^r_t))^2
\end{split}
\label{eq7}
\end{equation}
where $\theta_{V^{r}}$ and $\theta_{V^{l}}$ are the parameters of the value networks.

\begin{table*}[t]
\scriptsize
\centering
 \begin{tabular}{c|c|c|c|c|c|c|c|c}
\toprule
\multicolumn{3}{c|}{} & \multicolumn{3}{c|}{Charades-STA \cite{gao2017tall} }  & \multicolumn{3}{c}{ActivityNet \cite{krishna2017dense}}\\ \hline
Paradigm & Feature & Baseline   & IoU@0.7 & IoU@0.5 & MIoU & IoU@0.5 & IoU@0.3 & MIoU \\ \hline
\multirow{10}{*}{SL} &C3D &MCN \cite{anne2017localizing}, ICCV 2017 &4.44	&13.66 & 18.77& 10.17&22.07&15.99\\
 &C3D&CTRL \cite{gao2017tall}, ICCV 2017 &8.89	&23.63& -&14.36	&29.10&21.04\\
 &C3D&ACRN \cite{liu2018attentive}, SIGIR 2018 &9.65	&26.74 &26.97 &16.53	&31.75&24.49 \\
  &C3D&TGN \cite{chen2018temporally}, EMNLP 2018 &-&-&-&28.47	&45.51&- \\
 &C3D&MAC	\cite{ge2019mac}, WACV 2019 &12.23	&29.39& 29.01& - & - & -\\
 &C3D&SAP	\cite{chen2019semantic}, AAAI 2019 &13.36	&27.42& 28.56 & -& - & -  \\
 &C3D&QSPN \cite{xu2019multilevel}, AAAI 2019 &15.80	&35.60 & - &27.70	&45.30 & -  \\
 &C3D&ABLR \cite{yuan2018find}, AAAI 2019	&-&-&-&36.79	&55.67 &36.99 \\ \cline{2-9}
 &I3D&MAN \cite{zhang2019man}, CVPR 2019	&22.72&\textbf{46.53}&-&-&- &- \\
\midrule
\multirow{5}{*}{RL} &C3D &SM-RL \cite{wang2019language}, CVPR 2019 &11.17 &24.36 & 32.22 & -& - & -  \\
 &C3D &TripNet  \cite{hahn2019tripping}, CVPRW 2019 &14.50&36.61 &-&32.19 &48.42&-  \\
 &C3D &RWM \cite{he2019read}, AAAI 2019 &13.74	&34.12 &35.09 & 34.91 & 53.00 &36.25  \\
 &C3D &TSP-PRL (Ours) &17.69 &37.39&37.22&\textbf{38.76}&\textbf{56.08}&\textbf{39.21} \\ \cline{2-9}
 &Two-Stream &RWM \cite{he2019read}, AAAI 2019 &17.72	&37.23 &36.29 & - & - &-  \\
 &Two-Stream &TSP-PRL (Ours) &\textbf{24.73}	&45.30&\textbf{40.93}&-	&-&- \\
\bottomrule
\end{tabular}
\caption{The comparison performance (in \%) with state-of-the-art methods. The approaches in the first group are supervised learning (SL) based approaches and methods of the second group are reinforcement learning (RL) based approaches. ``-'' indicates that the corresponding values are not available.}
\label{table:result1}
 \end{table*}

Optimizing the root and leaf policies will simultaneously lead to the unstable training procedure. To avoid this, we design a progressive reinforcement learning (PRL) optimization procedure: for each set of $K$ iterations, PRL keeps one policy fixed and only trains the other policy. When reaching $K$ iterations, it switches the policy that is trained. The tree-structured policy based progressive reinforcement learning can be summarized as:
\begin{equation}
\psi = \lfloor \frac{i}{K} \rfloor  \bmod  2,
\label{eq8}
\end{equation}
\begin{equation}
\begin{split}
 \mathcal{L}_{tree} = & \mathcal{\psi} \times [\mathcal{L}_{root}(\theta_{\pi^{r}}) + \mathcal{L}_{root}(\theta_{V^{r}})] \\+
& (1-\mathcal{\psi}) \times [\mathcal{L}_{leaf}(\theta_{\pi^{l}}) + \mathcal{L}_{leaf}(\theta_{V^{l}})],
\end{split}
\label{eq9}
\end{equation}
where $\psi$ is a binary variable indicating the selection of the training policy. $i$ denotes the number of iterations in the entire training process. $\lfloor \rfloor $ is the lower bound integer of the division operation and $\bmod$ is the modulo function. These two policies promote each other mutually, as leaf policy provides accurate intrinsic rewards for root policy while the root policy selects the appropriate high-level semantic branch for further refinement of the leaf policy. The better leaf policy is, the more accurate intrinsic rewards will be provided. The more accurate the upper branch policy is selected, the better the leaf policy can be optimized. This progressive optimization ensures the agent to obtain a stable and outstanding performance in the RL setting. During testing, the tree-structured policy takes the best actions tuple $\left\langle a^r, a^l\right\rangle$ at each time step iteratively to obtain the final boundary.

\noindent\textbf{Alignment Network for \emph{Stop} Signal}. Traditional reinforcement learning approaches often include \emph{stop} signal as an additional action into the action space. Nevertheless, we design an alignment network to predict a confidence score $C_t$ for enabling the agent to have the idea of when to stop.
The optimization of the alignment network can be treated as an auxiliary supervision task since the temporal IoU can explicitly provide ground-truth information for confidence score. This network is optimized by minimizing the binary cross-entropy loss between $U_{t-1}$ and $C_t$:
\begin{footnotesize}
\begin{equation}
\mathcal{L}_{align}= \frac{1}{M}\sum_{m=1}^{M} \sum_{t=1}^{T_{max}} [U_{t-1} \log \sigma(C_t)+(1-U_{t-1}) \log (1-\sigma(C_t))].
\label{eq13}
\end{equation}
\end{footnotesize}
During testing, the agent will interact with the environment by $T_{max}$ steps and obtain a series of $C_t$.
Then the agent gets the maximum of $C_t$, which indicates that the alignment network considers $U_{t-1}$ has a maximal temporal IoU. So $t-1$ is the termination step.
The alignment network is optimized in the whole training procedure. The overall loss function in the proposed framework is summarized as:
\begin{equation}
 \mathcal{L} = \mathcal{L}_{tree} + \lambda \mathcal{L}_{align}.
\label{eq14}
\end{equation}
where $\lambda$ is a weighting parameter to achieve a tradeoff between two types of loss.

\section{Experiments}

\subsection{Datasets and Evaluation Metrics}
\noindent\textbf{Datasets}. The models are evaluated on two widely used datasets: Charades-STA \cite{gao2017tall} and ActivityNet \cite{krishna2017dense}.
Gao \emph{et al.} \cite{gao2017tall} extended the original Charades dataset \cite{sigurdsson2016hollywood} to generate sentence-clip annotations and created the Charades-STA dataset, which comprises 12,408 sentence-clip pairs for training, and 3,720 for testing. The average duration of the videos is 29.8 seconds and the described temporally annotated clips are 8 seconds long on average.
ActivityNet \cite{krishna2017dense} contains 37,421 and 17,505 video-sentence pairs for training and testing. The videos in ActivityNet are 2 minutes long on average and the described temporally annotated clips are 36 seconds long on average.
ActivityNet dataset is introduced to validate the robustness of the proposed algorithm toward longer and more diverse videos.

\noindent\textbf{Evaluation Metrics}. Following previous works \cite{gao2017tall,yuan2018find}, we adopt two metrics to evaluate the model for this task. ``IoU@ $\epsilon$'' means the percentage of the sentence queries which have temporal IoU larger than $\epsilon$. ``MIoU'' denotes the average IoU for all the sentence queries.

\subsection{Implementation Details}
The initial boundary is set to $L_{0} = [N/4; 3N/4]$, where $N$ denotes the clips numbers of the video. $N/4$ and $3N/4$ denote the start and end clip indices of the boundary respectively. The parameters $Z$ is set to 16 and $80$ respectively for Charades-STA and ActivityNet Datasets.
We utilize two mainstream structures of action classifiers (i.e., C3D \cite{tran2015learning} and Two-Stream \cite{wang2016temporal}) for video feature extraction on Charades-STA dataset. For ActivityNet, we merely employ C3D model to verify the general applicability of the proposed approach.
 The size of the hidden state in GRU is set to 1024. In the training stage of TSP-PRL, the batch size is set to 16 and the learning rate is 0.001 with Adam optimizer. The factor $\zeta$ is fixed to 1 in the reward settings. The hyper-parameters $\alpha$, $\gamma$ and $\lambda$ is fixed to 0.1, 0.4 and 1 receptively. For all experiments in this paper, we use $K$ = 200 in TSP-PRL. $T_{max}$ is set to $20$ to achieve the best trade off between accuracy and efficiency in the procedure of training and testing.
\begin{figure*}[t]
\centering
\subfigure[\scriptsize{``Ours w/o TSP-10'' with $T$ grows}]{
\includegraphics[width=0.238\linewidth]{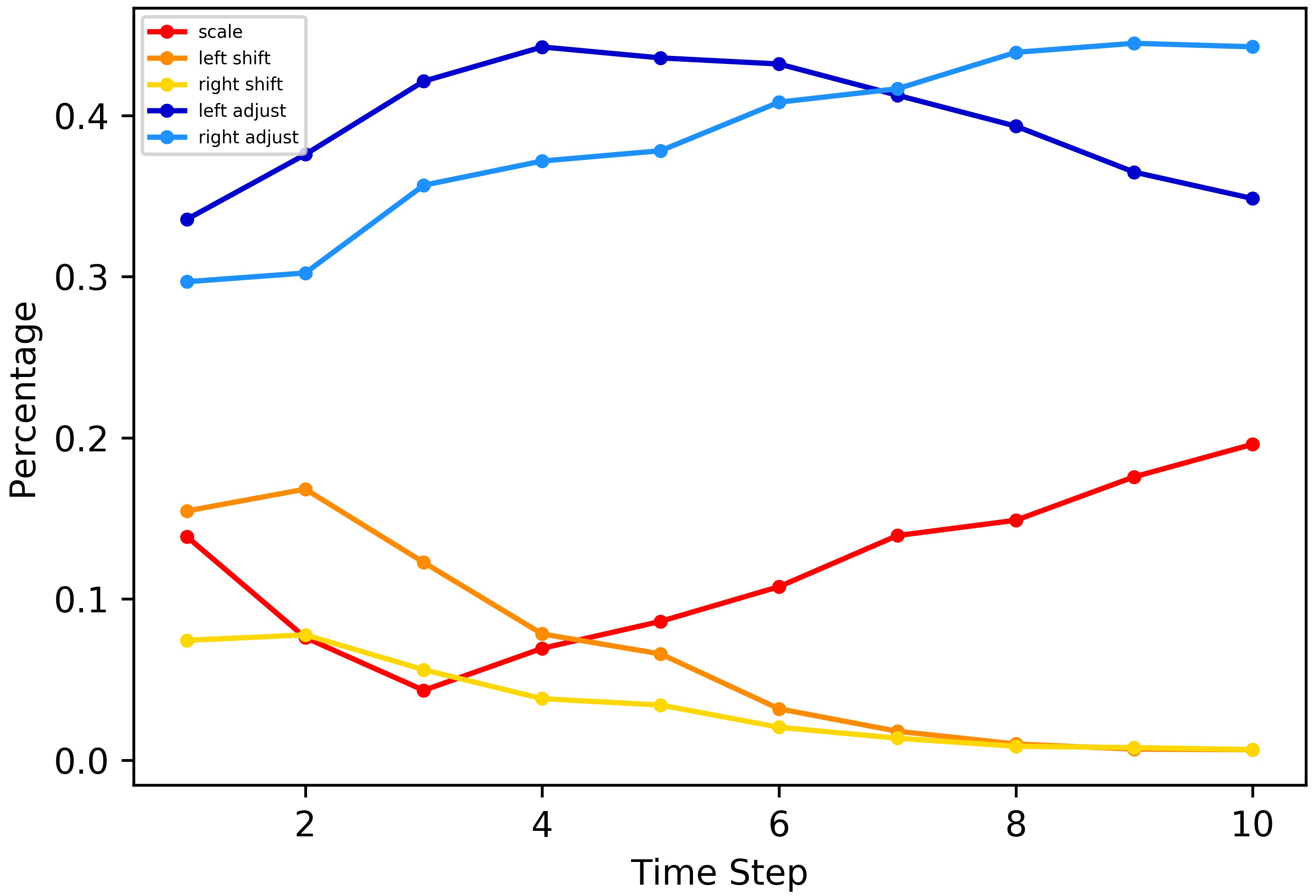}}
\subfigure[\scriptsize{``Ours-10'' with $T$ grows}]{
\includegraphics[width=0.238\linewidth]{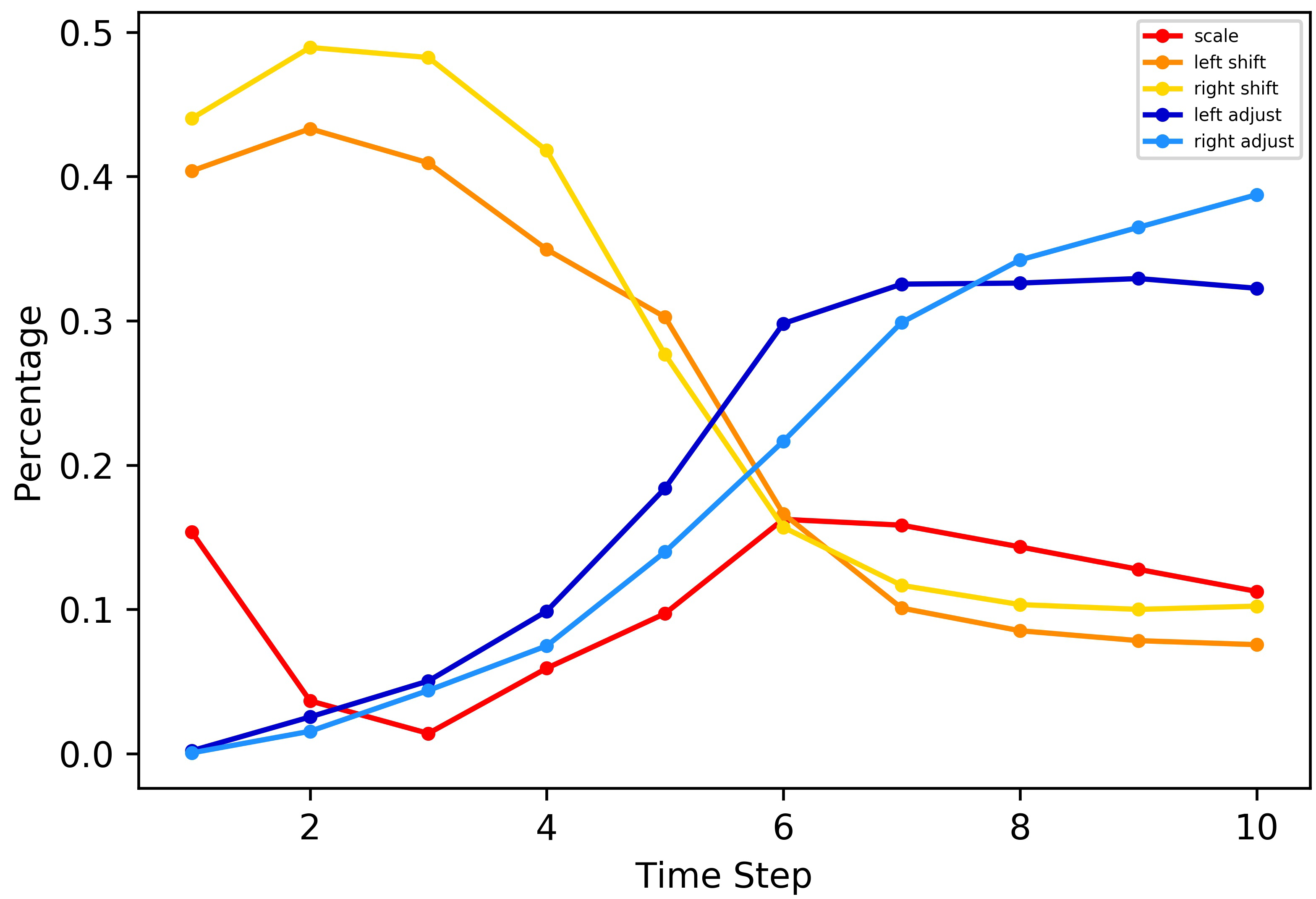}}
\subfigure[\scriptsize{``Ours w/o TSP-10'' with IoU increases}]{
\includegraphics[width=0.238\linewidth]{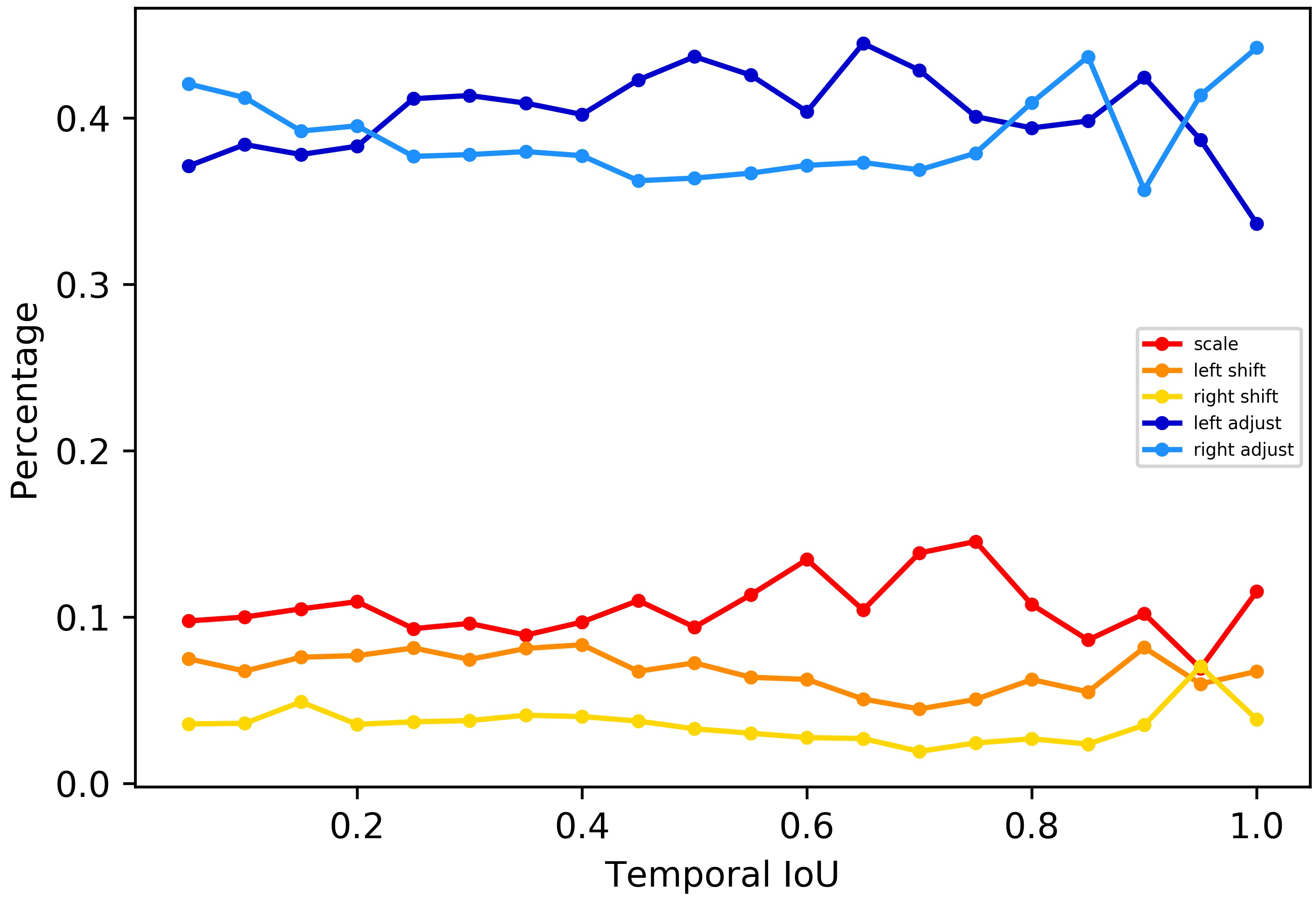}}
\subfigure[\scriptsize{``Ours-10'' with IoU increases}]{
\includegraphics[width=0.238\linewidth]{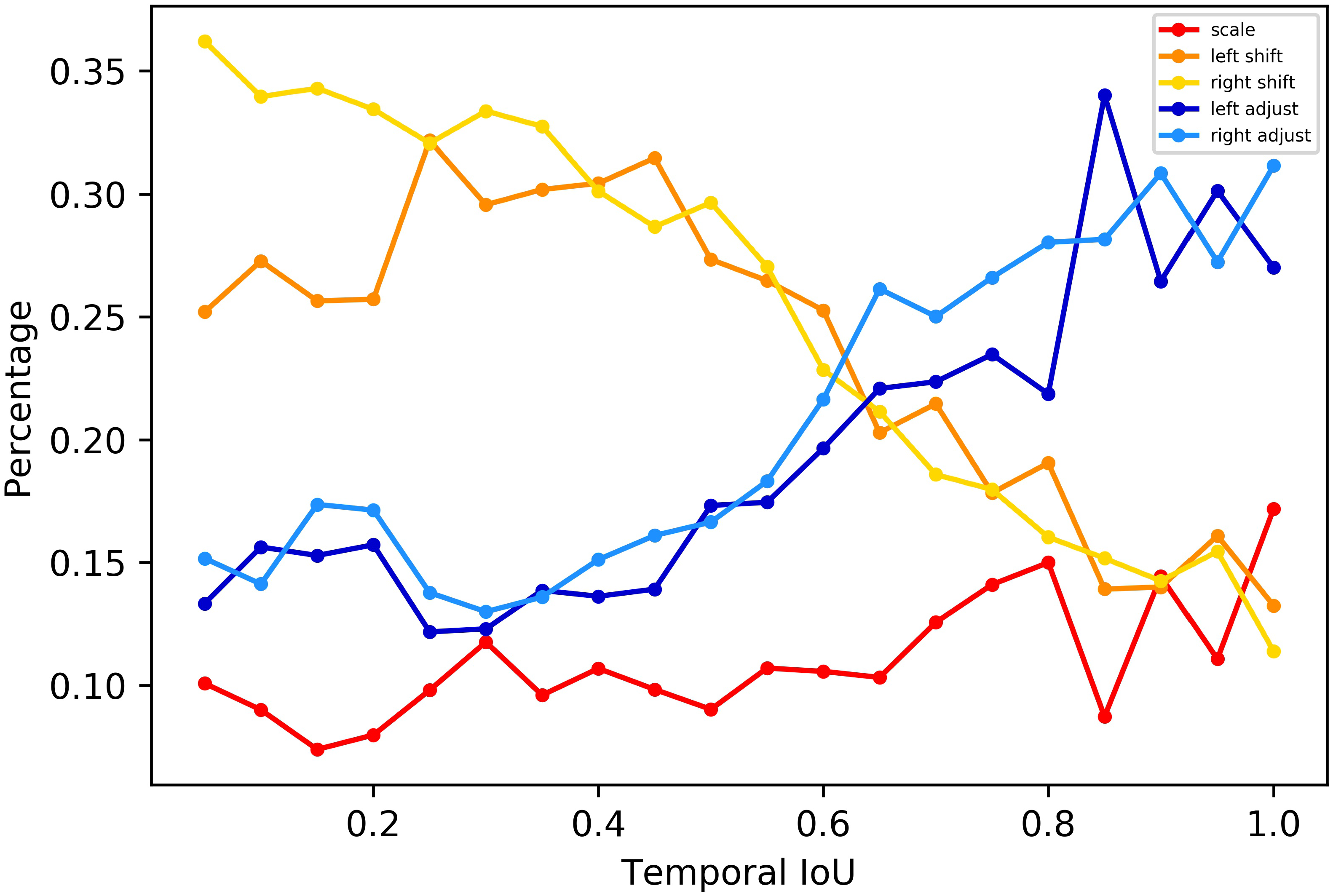}}
\caption{The proportion curve of the selected semantic branch as time step ($T$) grows and IoU increases. Correspondence between line color and semantic branch: 1) red : scale branch; 2) orange: left shift branch; 3) yellow: right shift branch; 4) dark blue: left adjust branch; 5) light blue: right adjust branch. Best viewed in color.}
\label{fig:proportion}
\end{figure*}

\subsection{Experimental Results}

\subsubsection{Comparison with the state-of-the-art algorithms.}

In this subsection, we compare TSP-PRL with 12 existing state-of-the-art methods on the Charades-STA and ActivityNet datasets in Table \ref{table:result1}. We re-implement ACRN \cite{liu2018attentive}, MAC \cite{ge2019mac} and RWM \cite{he2019read} and show their performance results in our experiments. The results of other approaches are taken from their paper.
The well-performing methods, such as QSPN \cite{xu2019multilevel}, ABLR \cite{yuan2018find} and MAN \cite{zhang2019man} all delve deep into the multi-modal features representation and fusion between the verbal and visual modalities.
Our approach focuses more on localization optimization, and it is complementary to the above-mentioned feature modeling methods actually.
On the one hand, TSP-PRL consistently outperforms these state-of-the-art methods, w.r.t all metrics with C3D feature.
For example, our method improves IoU@0.7 by 1.89\% compared with the previous best \cite{xu2019multilevel} on the Charades-STA.
For ActivityNet, the MIoU of TSP-PRL achieves the comparative enhancement over ABLR by 6.0\%.
MAN \cite{zhang2019man} employs stronger I3D \cite{carreira2017quo} to extract video features and obtain outstanding performance.
Our method with the Two-Stream feature manages to improve IoU@0.7 from 22.72\%  to 24,73\% on the Charades-STA.
On the other hand, TSP-PRL manages to obtain more flexible boundary, avoiding exhaustive sliding window searching compared with the supervised learning-based (SL) methods. SL methods are easy to suffer from overfitting and address this task like a black-box that lack of interpretability. While TSP-PRL contributes to achieving more efficient, impressive and heuristic grounding results.

\subsection{Ablative Study}
 As shown in Table \ref{table:result2}, we perform extensive ablation studies and demonstrate the effects of several essential  components in our framework. The Charades-STA dataset adopts the Two-stream based feature and the ActivityNet dataset uses the C3D based feature.
\begin{table}[t]
\centering
\scriptsize
 \begin{tabular}{c|c|c|c|c}
\toprule
Datasets & \multicolumn{2}{c|}{Charades-STA}  & \multicolumn{2}{c}{ActivityNet}\\ \hline
Metrics   &IoU@0.7 &IoU@0.5 &IoU@0.5 &IoU@0.3  \\ \hline
Ours w/o TSP-10 &17.13	&38.06 & 32.09&	49.35\\
Ours w/o TSP-20 &20.67	&41.31 & 34.39&	51.96\\
Ours w/o TSP-30 &22.40	&43.38 & 35.32&	52.77\\ \hline
Ours w/o IR &20.35	&40.64&35.03&52.64\\
Ours w/o ER &23.18	&44.41&37.20&55.78\\ \hline
Ours w/o AN &19.03  &39.78 & 33.89&  51.03\\ \hline
Ours-10 &22.85	&44.24 &37.53 &55.17 \\
Ours-20 &24.73	&45.30 &38.76 &56.08\\
Ours-30 &\textbf{24.75}	&\textbf{45.45} &\textbf{38.82} &\textbf{56.02} \\
\bottomrule
\end{tabular}
\caption{Comparison of the metrics (in \%) of the proposed approach and four variants of our approach. ``-$j$'' denotes that we set the max episode lengths to $j$ during testing. }
\label{table:result2}
 \end{table}

\noindent \textbf{Analysis of Tree-Structured Policy.}
To validate the significance of the tree-structured policy, we design the flat policy (denote as ``Ours w/o TSP'') that removes the tree-structured policy in our approach and directly maps state feature into a primitive action. As shown in Table \ref{table:result2}, flat policy declines IoU@0.7 to 17.13\%, 20.67\%, and 22.40\% at each level of $T_{max}$, with a decrease of 5.72\%, 4.06\%, and 2.35\% when compared with our approach.
Furthermore, it's performance suffers from a significant drop as $T_{max}$ decreases, which reveals that the flat policy relies heavily on the episode lengths to obtain better results. However, our approach manages to achieve outstanding performance with fewer steps.

In order to further explore whether the tree-structured policy can better perceive environment state and decompose complex policies, we summarize the proportion of the selected high-level semantic branch at each time step and IoU interval (0.05 for each interval). The percentage curves of two models (``Ours w/o TSP-10'' and ``Ours-10'') are depicted in Figure \ref{fig:proportion}. We can observe that the flat policy tends to choose the adjust based branches all the time and is not sensitive to the time step and IoU. However, our approach manages to select the shift based branches at first few steps to reduce the semantic gap faster. When the IoU increases or time step grows, the adjust based branches gradually dominant to regulate the boundary finely. Figure \ref{fig:proportion} clearly shows that tree-structured policy contributes to efficiently improving the ability to discover complex policies which can not be learned by flat policies.
To sum up, it is more intuitive and heuristic to employ the tree-structured policy, which can significantly reduce the search space and provide efficient and impressive grounding results. 

\noindent \textbf{Analysis of Root Reward.} To delve deep into the significance of each term in the root reward, we design two variants that simply remove the intrinsic reward item (denotes as ``Ours w/o IR'') and extrinsic reward item (denotes as ``Ours w/o ER'') in the definition of the root reward. As shown in Table \ref{table:result2}, removing the intrinsic reward term leads to an noticeable drop in performance. It indicates that the extrinsic reward item can not well reflect the quality of the root policy since this term is more relevant to the selected leaf policy.
``Ours w/o ER'' obtains 44.41\% and 37.20\% on IoU@0.5 on two datasets respectively, but it is still inferior to our approach.
Taking into account the direct impact (intrinsic reward) and indirect impact (extrinsic reward) simultaneously, our approach contributes to providing accurate credit assignment and obtaining a more impressive result.

\noindent \textbf{Analysis of \emph{Stop} Signal.} To demonstrate the effectiveness of the alignment network for \emph{stop} signal, we design a variant (denote as ``Ours w/o AN'') that removes the alignment network and directly includes the \emph{stop} signal as an additional action into the root policy.
The baseline assigns the agent a small negative reward in proportion with the step numbers.
As shown in Table \ref{table:result2}, ``Ours w/o AN'' gets a less prominent performance, which may be due to the fact that it is difficult to define an appropriate
reward function for the \emph{stop} signal in this task. However, our approach manages to learn the stop information with stronger supervision information via the alignment network, and it significantly increases the performance of all metrics by a large margin.

\section{Conclusions}
We formulate a novel Tree-Structured Policy based Progressive Reinforcement Learning (TSP-PRL) approach to address the task of temporally language grounding in untrimmed videos.
The tree-structured policy is invoked at each time step to reason a series of more robust primitive actions, which can sequentially regulate the temporal boundary via an iterative refinement process.
The tree-structured policy is optimized by a progressive reinforcement learning paradigm, which contributes to providing the task-oriented reward setting for correct credit assignment and optimizing the overall policy mutually and progressively.
Extensive experiments show that our approach achieves competitive performance over state-of-the-art methods on the widely used Charades-STA and ActivityNet datasets.

\small
\bibliography{AAAI-WuJ.682}
\bibliographystyle{aaai}
\end{document}